# Sim-to-Real Optimization of Complex Real World Mobile Network with Imperfect Information via Deep Reinforcement Learning from Self-play

Yongxi Tan[1*], Jin Yang[1], Xin Chen[2], Qitao Song[2], Yunjun Chen[2], Zhangxiang Ye[2], Zhenqiang Su[1]
[1] Futurewei Technologies Inc., Huawei R&D USA, USA
{yongxi.tan,jin.yang,zhenqiang.su}@huawei.com
[2] Huawei Technologies Co. Ltd, P.R.China
{chenxin,songqitao,chenyunjun,yezhangxiang}@huawei.com

**Abstract**

Mobile network that millions of people use every day is one of the most complex systems in the world. Optimization of mobile network to meet exploding customer demand and reduce capital/operation expenditures poses great challenges. Despite recent progress, application of deep reinforcement learning (DRL) to complex real world problem still remains unsolved, given data scarcity, partial observability, risk and complex rules/dynamics in real world, as well as the huge reality gap between simulation and real world. To bridge the reality gap, we introduce a Sim-to-Real framework to directly transfer learning from simulation to real world via graph convolutional neural network (CNN) — by abstracting partially observable mobile network into graph, then distilling domain-variant irregular graph into domain-invariant tensor in locally Euclidean space as input to CNN —, domain randomization and multi-task learning. We use a novel self-play mechanism to encourage competition among DRL agents for best record on multiple tasks via simulated annealing, just like athletes compete for world record in decathlon. We also propose a decentralized multi-agent, competitive and cooperative DRL method to coordinate the actions of multi-cells to maximize global reward and minimize negative impact to neighbor cells. Using 6 field trials on commercial mobile networks, we demonstrate for the first time that a DRL agent can successfully transfer learning from simulation to complex real world problem with imperfect information, complex rules/dynamics, huge state/action space, and multi-agent interactions, without any training in the real world.

## 1 Introduction

Despite the recent success of deep reinforcement learning (DRL) in games [1, 2, 3], application of DRL to complex real world problems still remains unsolved, due to data scarcity, imperfect information [4, 5] and complex rules/dynamics in the real world, as well as the big reality gap between simulation and real world [6, 7, 8]. In this work, we use DRL for one-shot optimization of complex real world mobile network that millions of people use every day (Figure 1). Coverage & capacity optimization (CCO) of mobile network poses great challenges [9]. First, mobile network is one of the most complex systems in real world since it is a multi-users, multi-cells, multi-technologies (3G, 4G, 5G) heterogeneous network. Second, it is critical to take actions simultaneously only once (one-shot, Figure 1) since CCO involves time-consuming and costly site visits to adjust vertical/horizontal angle (Tilt/Azimuth) of cell antenna. Third, mobile network is partially observable since important information about mobile network, environment and mobile users is often imperfect, missing or erroneous, e.g., unknown user location, missing information about building material and indoor map, wrong recorded value for Tilt/Azimuth. Fourth, state space is enormous, which depends on network parameters (e.g., number of cells, cell location/height, antenna Tilt/Azimuth/power, frequency band), environment (e.g., weather, terrain, tree, building location/3D shape/indoor map/material) and users (number of users, user location/distribution, moving speed/trajectory, mobile device). Fifth, action space is huge, e.g., 1.2e52 possible actions for 50 cells with 11 Tilt angle per cell. Finally, coordinating actions of multiple cells is crucial [10] since adjusting Tilt/Azimuth of a cell has significant impact to mobile users in this cell and neighbor cells. By analogy, imagine how difficult it is for three pilots (multi-cells) to try only once (one-shot) to land three airplanes (antennas) simultaneously at the same airport (mobile users) in dense fog (partial observability), with wrong instrument reading about current landing angle (Tilt and Azimuth).

## 2 Method

**Sim-to-Real DRL framework via graph-CNN**. To bridge the reality gap, first (Figure 2a), (a) we abstract the observations ($O$) for the partially observable mobile network into a graph, $G(V_c, V_u, E, A)=\Phi(O)$, with nodes being cells $V_c$ or users $V_u$, and weights of undirected edge $E$ or directed edge $A$ being certain relationships or interactions between nodes, e.g., distance between cells, signal strength received by a user equipment (UE) from a cell; (b) given the discrepancy between source (simulator) and target domain (real world) – e.g., 620 static UEs in simulator vs. millions of moving UEs in real world – we further distill the domain variant irregular subgraph around cell $Ci$ into domain-invariant 3D local tensor $P_{i,NN}$ as input to conventional CNN, via $\psi(G)$. Domain-invariant perception $P_{i,NN}$ is used as the same lens through which DRL agent views both simulated (during training) and real world mobile network (during inference) to bridge the reality gap.

Second (Figure 2b), if tasks in both source and target domain are similar, both source task distribution $\Omega_S$ and target task distribution $\Omega_T$ can be thought of drawn from the same task population $\Omega$. Therefore, to improve the generalization from source to target domain: (a) we design a wide range of diversified training/validation tasks in source domain, which most likely occur in target domain according to domain or prior knowledge; (b) we use domain randomization [7, 11-13] to further minimize the difference between $\Omega_S$ and $\Omega_T$ in the domain-invariant local perception space of $P_{i,NN}$ or latent space of CNN; (c) we use multi-task learning to leverage the commonalities shared by relevant tasks in $\Omega_S$ and $\Omega_T$ [14-17]; (d) we combine deep neural network with appropriate regularization techniques to prevent overfitting, e.g., L2 regularizer, batch normalization. General speaking, a DRL agent that has good performance on sufficient amount of diversified training/validation tasks in source domain should have better chance to achieve optimal performance on unseen but similar tasks in target domain. Ideally, we want to learn optimal policy $\pi_T^*$ in target domain to transit from initial state $S^0_T$ to optimal state $S^*_T = T_T(S^0_T, \pi_T^*)$ in one-shot, where $T_T$ is the state transition function in target domain. In practice, we instead learn optimal policy $\pi_S^*$ in source domain to approximate $\pi_T^*$, and further approximate $\pi_S^*$ by a neural network $\pi_S^\theta(P_{i,NN})$ with parameters $\theta$: $\pi_S^\theta \approx \pi_S^* \approx \pi_T^*$.

**Distill domain-variant irregular graph into domain-invariant local tensor**. Application of CNN to irregular data such as graph poses big challenges. To generalize conventional CNN from grid-based image to irregular graph, convolutional filters have recently been adapted to graph data, either in spatial domain by extracting local connected region from graph, or in spectral domain by using the spectrum of the graph Laplacian [18-20]. However, given the huge scale of real world network graph (e.g., billions of nodes) and the reality gap between simulation and real world, we take the opposite approach, i.e., adapting the graph data to conventional CNN by transforming domain-variant irregular graph to domain-invariant 3D tensor in locally Euclidean space as input to conventional CNN.

As illustrated in Figure 2a, for any top-K cell $C_i$ selected for adjustment, first, we calculate the affinity between $C_i$ and all other cell $N_j$ according to, e.g., geometry, overall interference; second, we select $N$ (=31) most important neighbor cells to be included in the field of view of $C_i$ based on affinity rank; third, we construct the grid-like local Euclidean coordinate from subgraph around $C_i$ by putting $C_i$ in the center and $N_j$ beside $C_i$ orderly along X/Y axis of $P_{i,NN}$ according to affinity; fourth, we distill multi-type relationships or interactions among cells and users in the subgraph around $C_i$ into $M$ (=52) channels along Z-axis; finally, CNN-based policy network takes local tensor $P_{i,NN}$ as input and output action $a^i$ for $C_i$ based on its own partial view: $\pi^\theta(a^i|P_{i,NN})$.

**Self-play to compete for best record**. Inspired by AlphaGo, we introduce a novel self-play mechanism to encourage competition for best record in history on multi-tasks, just like athletes compete for world record in decathlon. As illustrated in Figure 3, for any initial state $S^i_0$ of task $T_i$ drawn from source task distribution $\Omega_S$, if new actions $<a^1,a^2,\ldots,a^n>$ for cell $C_1,C_2,\ldots,C_n$ achieve better immediate global reward over all cells, $R_{new}$, than the best record $R_{best}$ in history for the same $S^i_0$ of the same $T_i$ by a threshold: $\Delta R_g=R_{new}-R_{best}>=Th_{ge}$, we encourage these actions by backpropagating a gradient $g_e=T_e(R-B)\nabla_\theta \log \pi^\theta(a^i|P_{i,NN})$ to policy network $\pi^\theta(a^i|P_{i,NN})$ for each action $a^i$ of cell $C_i$ individually, where $T_e$ is a transformation function (e.g., +1*Abs(x), where Abs is absolute value function), $R$ is expected total reward, $B$ is baseline in REINFORCE [21]. If $\Delta R_g<=Th_{gp}$, we penalize these actions by gradient $g_p=T_p(R-B)\nabla_\theta \log \pi^\theta(a^i|P_{i,NN})$, where $T_p$ is a transformation function, e.g, -1*Abs. If $Th_{gp}<\Delta R_g <Th_{ge}$, we use simulated annealing (SA) to make decision to avoid local optimum and approximate global optimum, by comparing an uniform random number $\in [0,1]$ with global acceptance probability $P_g=1/(1+\exp(\Delta R_g/T_g))$, where $T_g$ is global SA temperature annealed according to certain cooling schedule.

**Decentralized multi-agent self-play competitive and cooperative DRL (S2C)**. Self-play mechanism is good at improving the global reward over all cells by competition. However, this approach has potential problems for decentralized multi-agent partial observable Markov decision process [22, 23]. As illustrated in Figure 3b — given that each top-K cell $C_i$ takes action $a^i$ based on its own partial view $P_{i,NN}$ around $C_i$, whereas global reward is calculated over all cells after one-shot CCO — it is possible that local cell reward for $C_i$ or its neighbor cells may be negative due to the bad action of $C_i$, even if the global reward is better than the best record. Therefore it is crucial to coordinate the actions taken for multiple cells in the mobile network.

We propose a decentralized multi-agent self-play competitive and cooperative DRL method (S2C) to coordinate the actions of multi-cells to not only compete as a team for best global reward via self-play, but also cooperate with each other to minimize negative impact to neighbor cells. As illustrated in Figure 3, each cell/agent takes action by its own local perception $P_{i,NN}$ using the same policy network $\pi^\theta(a^i|P_{i,NN})$. If new actions $<a^1,a^2,\ldots,a^n>$ are rejected at global level due to either $\Delta R_g<=Th_{gp}$ or global acceptance probability $P_g$, we penalize these actions by gradient $g_p$ for each action $a^i$ individually. Otherwise, for each action $a^i$, if the local reward $R_{Ci}$ for cell $C_i$ is larger than a threshold $R_{Ci}>=Th_{ce}$, then we accept action $a^i$ for $C_i$ with gradient $g_e$; if $R_{Ci}<=Th_{cp}$, then we reject it with gradient $g_p$; if $Th_{cp}< R_{Ci} <Th_{ce}$, we use SA to decide if accepting action $a^i$ for $C_i$ with cell level acceptance probability $P_c=1/(1+\exp(R_{Ci}/T_c))$, where $T_c$ is cell level SA temperature annealed according to certain cooling schedule.

## 3 Experiments and results

**Supervised learning of policy network**. First, we generated 2.38 millions CCO tasks $T_i$ (238 mobile networks, 10,000 random Tilt settings per network, Figure 4a) in Netlab simulator for training and validation. Second, we design a SA algorithm to optimize the CCO task sequentially in 10 iterations (10 shots CCO). Then we use it to generate labeled data – including $P^0_{i,NN}$ for each top-K cell $C_i$ at initial state $S^0_k$, and corresponding best Tilt action $a^i_{best} \in [-5°,5°]$ for $C_i$ in 10 shots – from 10k training tasks (500 initial states × 20 mobile networks). Supervised learning policy network (SL-DNN) is a residual network [24] with 32×32×52 tensor $P^0_{i,NN}$ as input and 11 outputs for labeled Tilt action $a^i_{best}$. Using 146k training data, we achieved accuracy of 78.4% (for Tilt difference between predicted and labeled <=1°) and 91.5% (for Tilt difference <=2°) for 16k hold-out test data.

**Multi-task deep reinforcement learning in Netlab simulator**. As shown in Figure 5, S2C achieved better performance than DQN [1], A3C [25], Double Q [26], supervised learning (SL-DNN) and simulated annealing based 10-shot CCO (SA 10-Shots), in terms of global reward averaged over all validation tasks (AvgGlobalReward, 6.46% for S2C), and ratio of validation tasks with positive global reward to all validation tasks (RatioPositiveReward, 94% for S2C). For comparison purpose, we also trained another supervised learning policy network (SL-DNN-2) using 874k training data (generated by SA 10-shots for 71,400 tasks), with prediction accuracy of 83% (for Tilt difference <= 1°) and 94% (for Tilt difference <=2°) for 65k hold-out test data. The validation result of SL-DNN-2 was not as good as other DRL models, even if it has pretty good prediction accuracy for the Tilt actions. This is because the objective of SL-DNN-2 is to maximize prediction accuracy of best action from SA 10-shots, rather than maximize the global reward for task.

**Within-simulator and cross-simulator transfer learning**. To evaluate the within-simulator generalization, we tested the same S2C model on 5481 unseen test tasks over 238 simulated mobile networks in Netlab, without any further training. As shown in Figure 6a, S2C achieved good within-simulator generalization, with AvgGlobalReward dropped by only 0.86% to 5.60% and RatioPositiveReward dropped by only 2% to 92%. To evaluate the cross-simulator generalization, we further tested the same S2C model on 7693 unseen test tasks over 160 unseen mobile networks in another simulator (Unet), without any further training. Unet is a good candidate for evaluation of cross-domain transfer learning, since it is quite different from Netlab: mobile networks simulated in Unet are much larger (100-140 cells and 2480-19840 static UEs per network, Figure 4b) than that in Netlab (Figure 4a); different radio propagation models and indoor penetration loss are simulated in Unet. As illustrated in Figure 6b, good results were also achieved with 4.93% AvgGlobalReward and 95.7% RatioPositiveReward.

**Direct transfer learning from simulation to real world mobile network.** To evaluate the generalization from simulation to real world without any training in real world, we performed 6 field trials on 4 commercial mobile networks that have never been simulated in both simulators before, and are very different from all simulated mobile networks. For example, vertical or horizontal multi-frequency network (MFN) or carrier aggregations (CA) has never been simulated in both simulators before; user distribution/number in real world mobile network is temporal-spatial dynamic and very different from static UE distribution/number in simulators (e.g., hundreds to thousands of static UEs in simulation, and millions of moving UEs in real world); real world mobile networks have very different cell/building layouts, indoor map, building materials and radio propagation. We separated commercial mobile network A into 2 neighboring clusters $C_1$/$C_2$ (66/47 cells, vertical MFN), and performed one trial for each cluster, with 2.03% RSRP (Reference Signals Received Power, coverage indicator) and 5.62% RSRQ (Reference Signal Received Quality, interference/capacity indicator) improvement in $C_1$, as well as 3.17% RSRP and 4.86% RSRQ improvement for $C_2$. The 3rd trial was done for whole mobile network A ($C_1$+$C_2$, 113 cells, vertical MFN), and no significant improvement was observed since most gain has been achieved in first 2 trials. In 4th trial on mobile network B (151 cells, vertical MFN), we achieved 10.79% RSRP and 6.74% RSRQ improvement. In 5th trial on mobile network C (131 cells, horizontal MFN and CA), no significant improvement was observed due to either little room for optimization or significant difference between mobile network C in real world and task distributions used in simulation. In 6th trial on commercial mobile network D (159 cells, horizontal MFN and CA), we achieved 9.55% RSRP and 12.42% RSRQ improvement.

# References

[1] Mnih, V., Kavukcuoglu, K., Silver, D., Rusu, A.A., Veness, J., Bellemare, M.G., Graves, A., Riedmiller, M., Fidjeland, A.K., Ostrovski, G., Petersen, S., Beattie, C., Sadik, A., Antonoglou, I., King, H., Kumaran, D., Daan Wierstra, Legg, S., & Hassabis, D. (2015) Human-level control through deep reinforcement learning. *Nature* **518**:529–533.

[2] Silver, D., Huang, A., Maddison, C.J., Guez, A., Sifre, L., van den Driessche, G., Schrittwieser, J., Antonoglou, I., Panneershelvam, V., Lanctot, M., Dieleman, S., Grewe, D., Nham, J., Kalchbrenner, N., Sutskever, I., Lillicrap, T., Leach, M., Kavukcuoglu, K., Graepel, T. & Hassabis, D. (2016) Mastering the game of Go with deep neural networks and tree search. *Nature* **529**:484–489.

[3] Silver, D., Schrittwieser, J., Simonyan, K, Antonoglou, I., Huang, A., Guez, A., Hubert, T., Baker, L., Lai, M., Bolton, A., Chen, Y., Lillicrap, T., Hui, F., Sifre, L., van den Driessche, G., Graepel, T. & Hassabis, D. (2017) Mastering the game of Go without human knowledge. *Nature* **550**:354–359

[4] Moravčík, M., Schmid, M., Burch, N., Lisý, V., Morrill, D., Bard, N., Davis, T., Waugh, K., Johanson, M. & Bowling, M. (2017) DeepStack: Expert-Level Artificial Intelligence in No-Limit Poker. *Science* **356**:508-513

[5] Brown, N. & Sandholm T. (2018) Superhuman AI for heads-up no-limit poker: Libratus beats top professionals. *Science* **359**:418-424.

[6] Rusu, A.A., Vecerik, M., Rothörl, R., Heess, N., Pascanu, R. & Hadsell, R. (2016) Sim-to-Real Robot Learning from Pixels with Progressive Nets. *Preprint* at https://arxiv.org/abs/1610.04286

[7] Tobin, J., Fong, R., Ray, A., Schneider, J., Zaremba, W. & Abbeel, P. (2017) Domain Randomization for Transferring Deep Neural Networks from Simulation to the Real World. *IROS* Preprint at https://arxiv.org/abs/1703.06907

[8] Bousmalis, K. & Levine, S. (2017) Closing the Simulation-to-Reality Gap for Deep Robotic Learning. *Google Research Blog* https://research.googleblog.com/2017/10/closing-simulation-to-reality-gap-for.html

[9] Fan, S., Tian, H. & Sengul, C. (2014) Self-optimization of coverage and capacity based on a fuzzy neural network with cooperative reinforcement learning. *EURASIP Journal on Wireless Communications and Networking* 2014:57.

[10] Vinyals, O., Ewalds, T., Bartunov, S., Georgiev, P., Vezhnevets, A.S., Yeo, M., Makhzani, A., Küttler, H., Agapiou, J., Schrittwieser, J., Quan, J., Gaffney, S., Petersen, S., Simonyan, K., Schaul, T., van Hasselt, H., Silver, D., Lillicrap, T., Calderone, K., Keet, P., Brunasso, A., Lawrence, D., Ekermo, A., Repp, J. & Tsing, R. (2017) StarCraft II: A New Challenge for Reinforcement Learning. *Preprint* at https://arxiv.org/abs/1708.04782

[11] Sadeghi, F., & Levine, S. (2016) CAD2RL: Real Single-Image Flight without a Single Real Image. *Preprint* at https://arxiv.org/abs/1611.04201

[12] Peng, X.B., Andrychowicz, M., Zaremba, W. & Abbeel, P. (2017) Sim-to-Real Transfer of Robotic Control with Dynamics Randomization. *Preprint* at https://arxiv.org/abs/1710.06537

[13] James, S., Davison, A.J., Johns, E. (2017) Transferring End-to-End Visuomotor Control from Simulation to Real World for a Multi-Stage Task. 1st Conference on Robot Learning (CoRL), *Preprint* at https://arxiv.org/abs/1707.02267

[14] Parisotto, E., Ba, J.L. & Salakhutdinov, R. (2016) Actor-Mimic: Deep Multitask and Transfer Reinforcement Learning. *ICLR* 2016. Preprint at https://arxiv.org/abs/1511.06342

[15] Ruder, S. (2017) An Overview of Multi-Task Learning in Deep Neural Networks. *Preprint* at https://arxiv.org/abs/1706.05098

[16] Caruana, R. (1997) Multi-task learning. Machine Learning. 28:41–75

[17] Zhang, Y., Yang, Q. (2017) Survey on Multi-Task Learning. Preprint at https://arxiv.org/abs/1707.08114

[18] Bronstein, M.M., Bruna, J., LeCun, Y., Szlam, A., Vandergheynst, P. (2017) Geometric deep learning: going beyond Euclidean data. IEEE Signal Processing Magazine, 34(4):18-42

[19] Bruna, J., Zaremba, W., Szlam, A., LeCun, Y. (2013) Spectral Networks and Locally Connected Networks on Graphs. 2013. Preprint at https://arxiv.org/abs/1312.6203

[20] Niepert, M., Ahmed, M., Kutzkov, K. (2016) Learning Convolutional Neural Networks for Graphs. ICML 2016. Preprint at https://arxiv.org/abs/1605.05273

[21] Williams, R.J. (1992) Simple Statistical Gradient-Following Algorithm for Connectionist Reinforcement Learning *Machine Learning* **8**:229–256

[22] Bernstein, D.S., Givan, R., Immerman, N., Zilberstein, S. (2002) The Complexity of Decentralized Control of Markov Decision Processes. Mathematics of Operations Research archive, 27: 819-840

[23] Oliehoek, F.A., Amato, C. (2016) A Concise Introduction to Decentralized POMDPs. Springer

[24] He, K., Zhang, X., Ren, S. & Sun, J. (2015) Deep Residual Learning for Image Recognition. *Preprint* at https://arxiv.org/abs/1512.03385

[25] Mnih, V., Badia, A.P., Mirza, M., Graves, A., Lillicrap, T.P., Harley, T., Silver, D. & Kavukcuoglu, K. (2016) Asynchronous Methods for Deep Reinforcement Learning. *Preprint* at https://arxiv.org/abs/1602.01783

[26] van Hasselt, H., Guez, A. & Silver, D. (2016) Deep Reinforcement Learning with Double Q-learning. *AAAI'16 Proceedings of the Thirtieth AAAI Conference on Artificial Intelligence* 2094-2100.

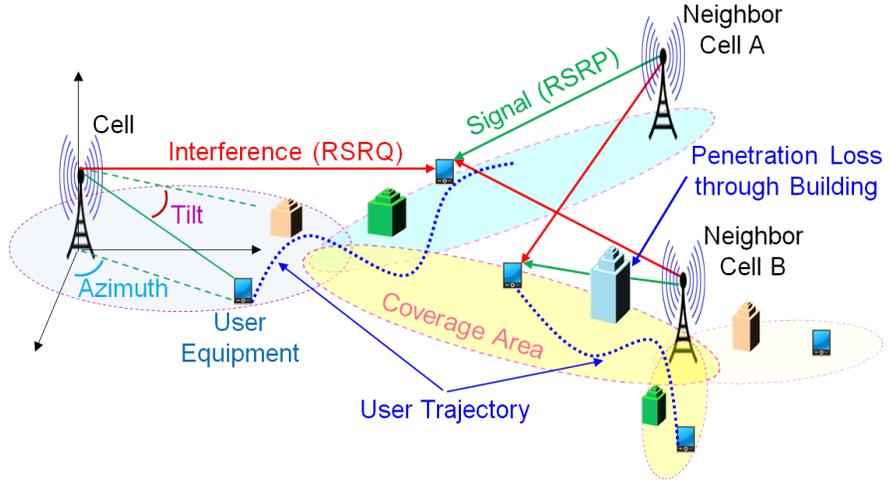

(a)

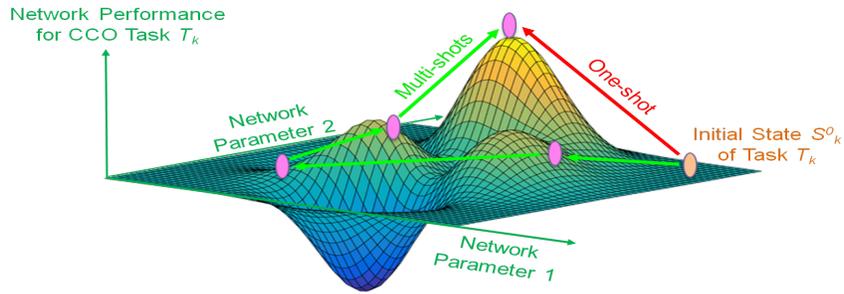

(b)

Figure 1: One-shot coverage and capacity optimization (CCO) of partially observable mobile network. (a) CCO task: adjusting actions (e.g., Tilt, Azimuth) of multiple cells to improve signal coverage (RSRP – reference signal received power) and reduce interference (RSRQ - reference signal received quality) that are received by all user equipments (UEs) inside mobile network. Temporal and spatial observations — e.g., measurement report (MR) for RSRP and RSRQ — depend on the state of partially observable mobile network, including wireless network parameters, environment, and users. (b) One-shot vs. multi-shot optimization.

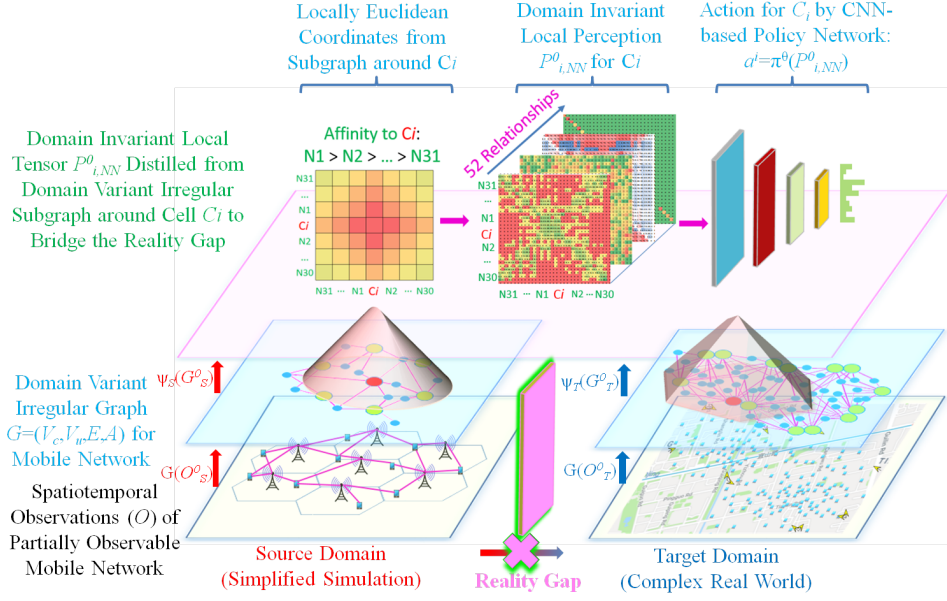

(a)

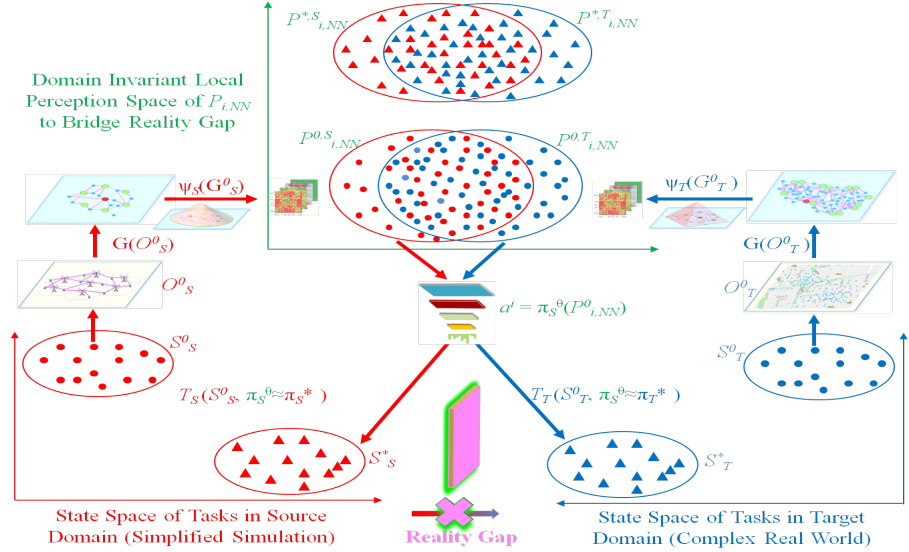

(b)

Figure 2: Sim-to-Real DRL framework. (a) Graph-CNN: first, we abstracts partial observations of interactions between cells and users into a domain-variant irregular network graph $G = (V_c, V_u, E, A)$; second, field of view (FOV) of cell $Ci$ (red node) is determined according to affinity between $Ci$ and all other cells; third, domain-invariant local tensor $P_{i,NN}$ for $Ci$ is distilled from 52 relationships between nodes in the subgraph within FOV of $Ci$; finally, CNN-based policy network takes $P_{i,NN}$ as input to output action $a^i$ for $Ci$. (b) Improve generalization via source task design, domain randomization and multi-task learning.

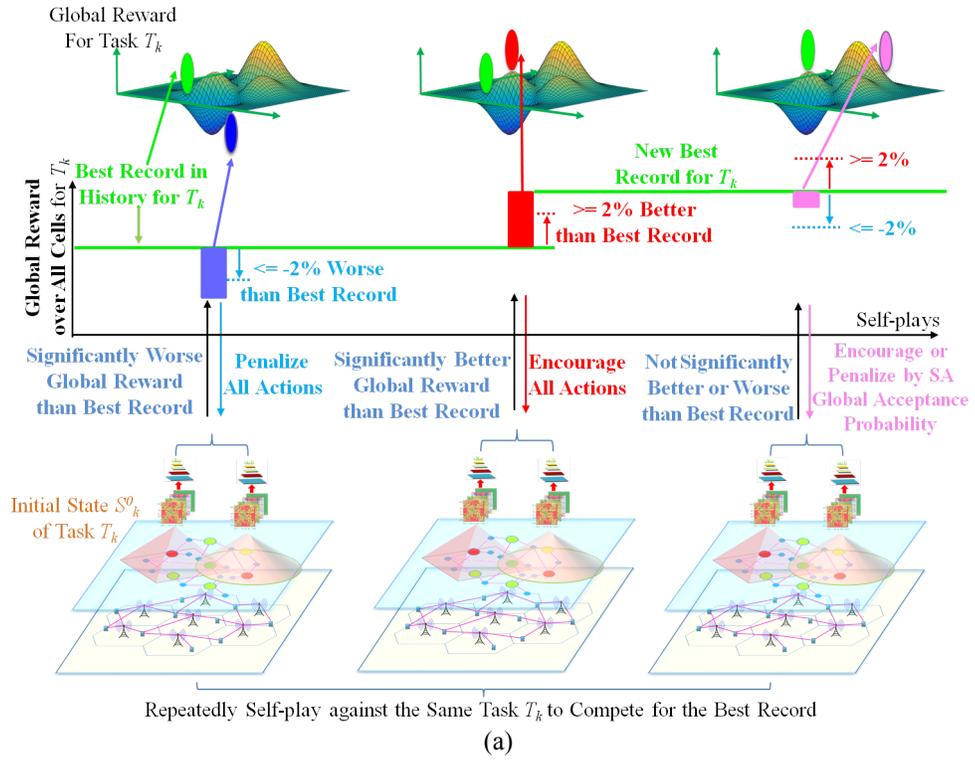

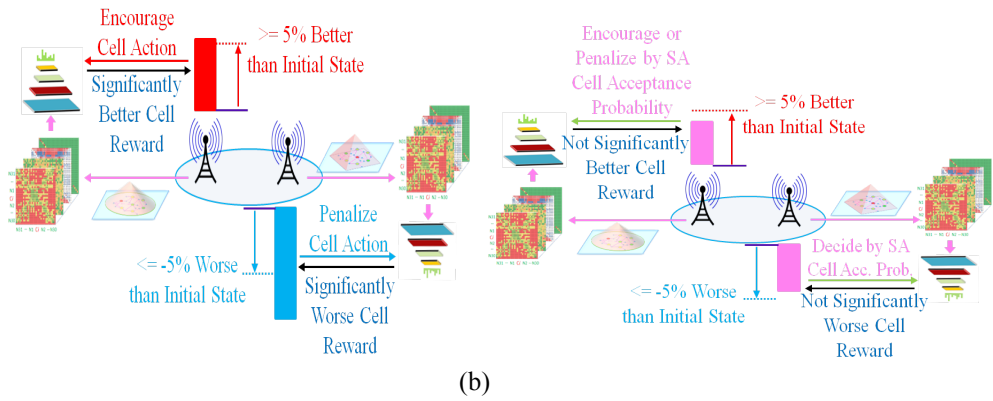

Figure 3: (a) Self-play to compete for best record on the same task. (b) Multi-cell cooperation.

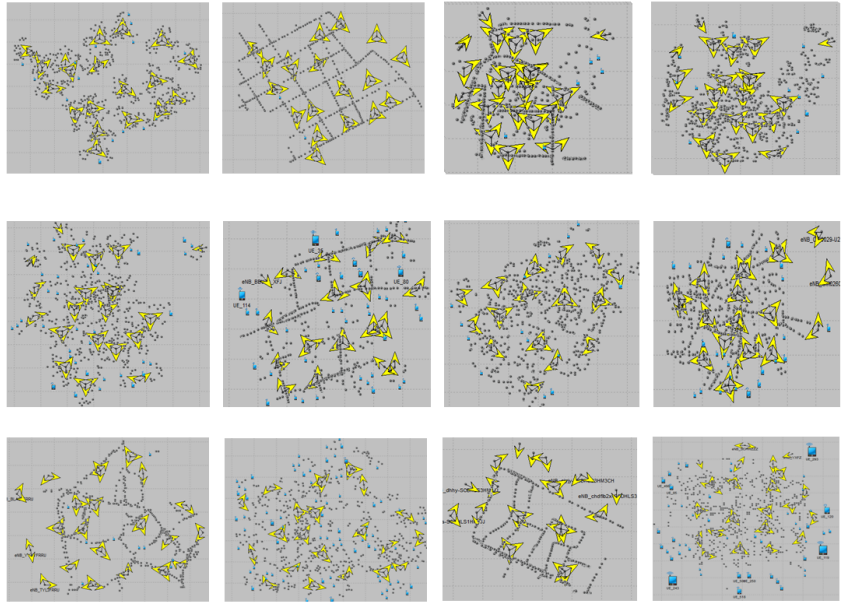

(a)

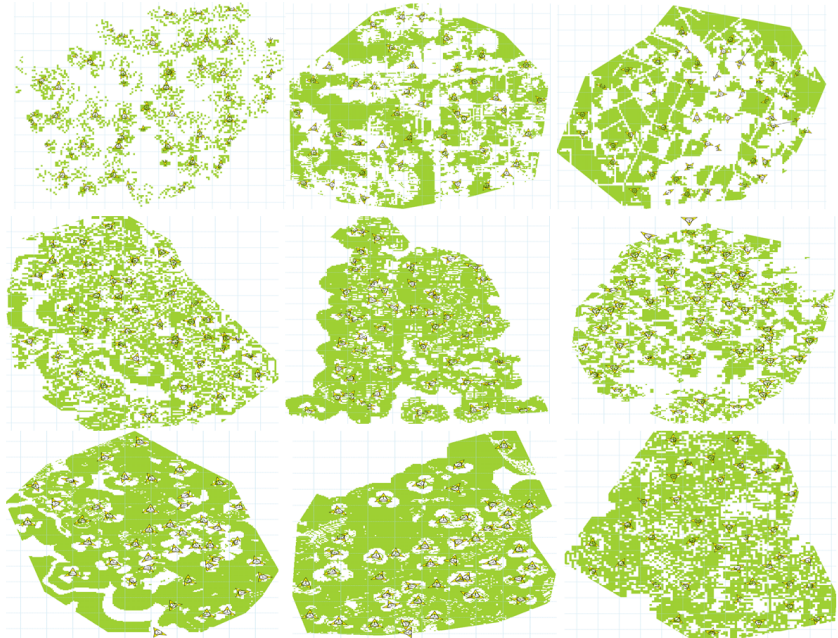

(b)

Figure 4: (a) 12 exemplary mobile networks simulated in Netlab simulator (yellow arrow denotes cell, blue or black dot denotes user, 30-60 cells and 400-620 static user per mobile network). (b) 9 exemplary mobile networks simulated in Unet simulator (yellow arrow denotes cell, green dot denotes user; 100-140 cells and 2480-19840 static users per mobile network).

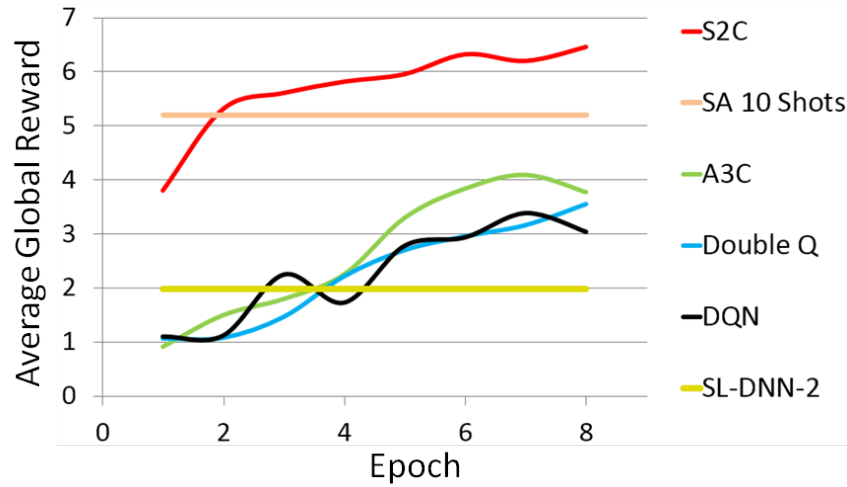

(a)

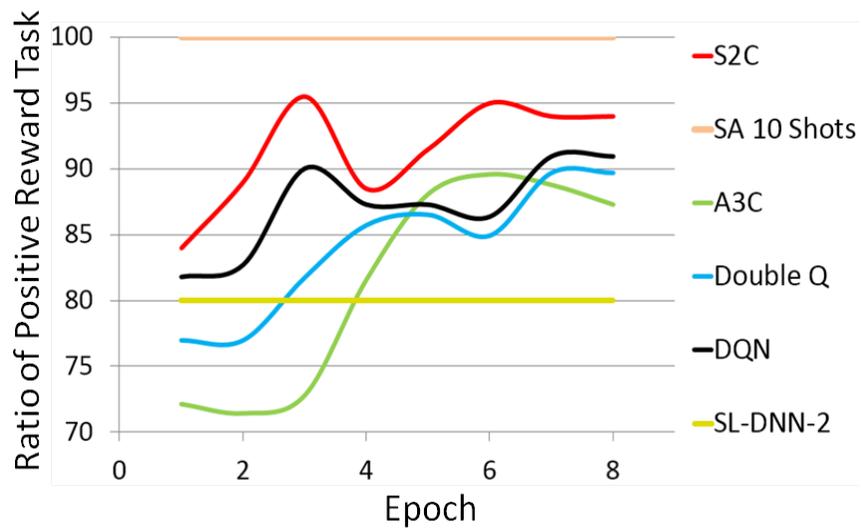

(b)

Figure 5: Validation result. (a) Global reward averaged over all validation tasks; (b) Ratio of validation tasks with positive global reward to all validation tasks.

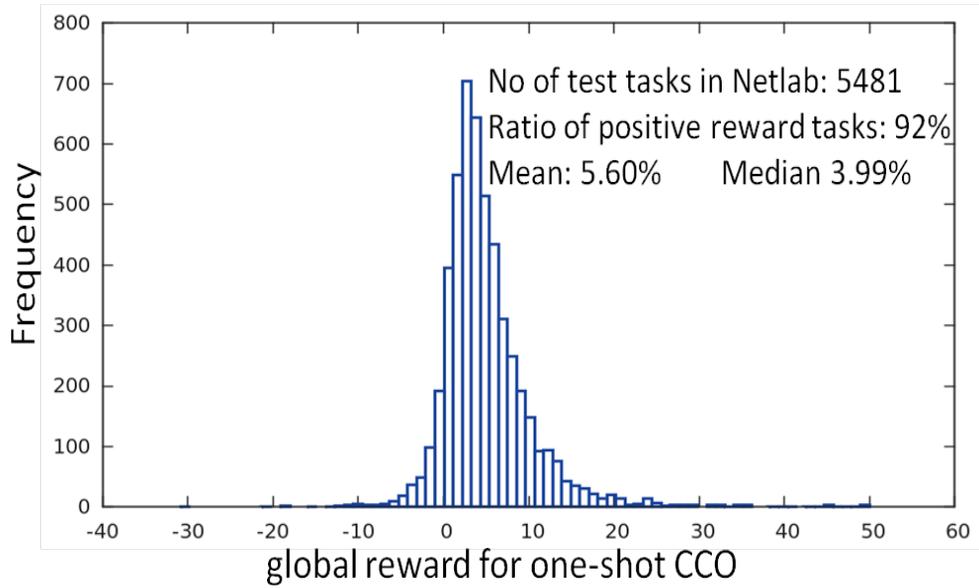

(a)

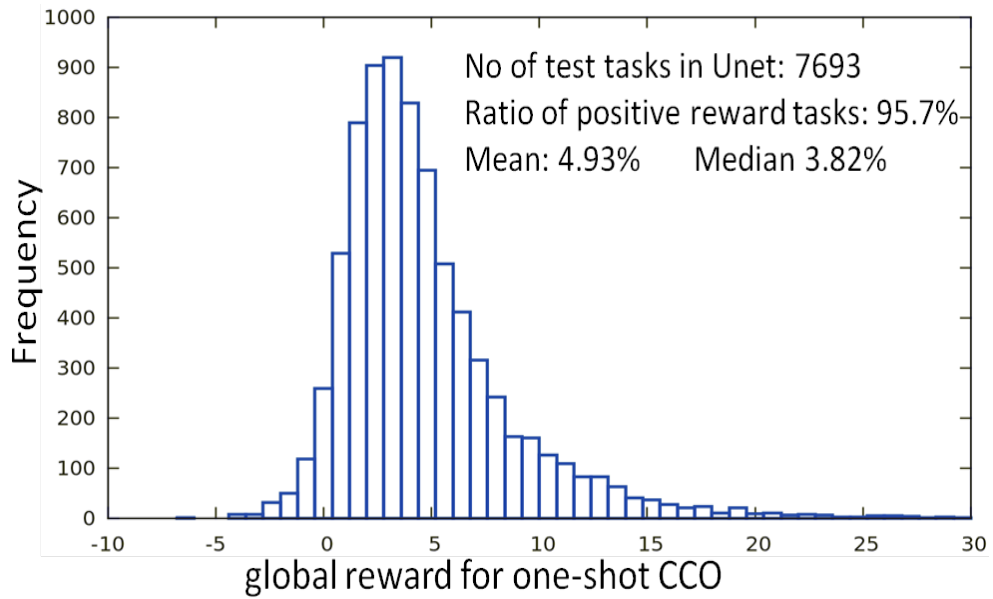

(b)

Figure 6: (a) Within-simulator transfer learning: tested against 5481 unseen test tasks over 238 simulated mobile networks in Netlab; (b) Cross-simulator transfer learning from Netlab simulator to Unet simulator: tested against 7693 unseen test tasks over 160 unseen mobile networks in Unet.

12